\documentclass{article}

\usepackage{arxiv}

\usepackage[utf8]{inputenc} 
\usepackage[T1]{fontenc}    
\usepackage{hyperref}       
\usepackage{url}            
\usepackage{booktabs}       
\usepackage{amsfonts}       
\usepackage{nicefrac}       
\usepackage{microtype}      
\usepackage{lipsum}		
\usepackage{graphicx}
\usepackage{natbib}
\usepackage{doi}

\title{Evaluating Generative AI-Enhanced Content: A Conceptual Framework Using Qualitative, Quantitative, and Mixed-Methods Approaches}

\author{ \href{https://orcid.org/0000-0003-3314-4281}{\includegraphics[scale=0.06]{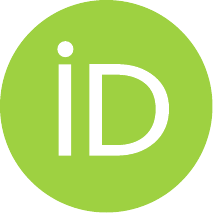}\hspace{1mm}Saman Sarraf}\thanks{Dr. Saman Sarraf is a Senior Member of IEEE (SMIEEE)} \\
	Santa Clara Valley Section, Institute of Electrical and Electronics Engineers\\
	Santa Clara, CA 94085, USA\\
	\texttt{samansarraf@ieee.org} \\
}

\renewcommand{\shorttitle}{Technical Report}

\hypersetup{
pdftitle={Evaluating Generative AI-Enhanced Content: A Conceptual Framework Using Qualitative, Quantitative, and Mixed-Methods Approaches},
pdfsubject={q-bio.NC, q-bio.QM},
pdfauthor={Saman Sarraf},
pdfkeywords={GenAI, Qualitative, Quantitative},
}

\begin{document}
\maketitle

\begin{abstract}
Generative AI (GenAI) has revolutionized content generation, offering transformative capabilities for improving language coherence, readability, and overall quality. This manuscript explores the application of qualitative, quantitative, and mixed-methods research approaches to evaluate the performance of GenAI models in enhancing scientific writing. Using a hypothetical use case involving a collaborative medical imaging manuscript, we demonstrate how each method provides unique insights into the impact of GenAI. Qualitative methods gather in-depth feedback from expert reviewers, analyzing their responses using thematic analysis tools to capture nuanced improvements and identify limitations. Quantitative approaches employ automated metrics such as BLEU, ROUGE, and readability scores, as well as user surveys, to objectively measure improvements in coherence, fluency, and structure. Mixed-methods research integrates these strengths, combining statistical evaluations with detailed qualitative insights to provide a comprehensive assessment. These research methods enable quantifying improvement levels in GenAI-generated content, addressing critical aspects of linguistic quality and technical accuracy. They also offer a robust framework for benchmarking GenAI tools against traditional editing processes, ensuring the reliability and effectiveness of these technologies. By leveraging these methodologies, researchers can evaluate the performance boost driven by GenAI, refine its applications, and guide its responsible adoption in high-stakes domains like healthcare and scientific research. This work underscores the importance of rigorous evaluation frameworks for advancing trust and innovation in GenAI.
\end{abstract}


\section{Introduction}

Natural Language Processing (NLP) is a cornerstone of artificial intelligence (AI) that focuses on enabling machines to understand, interpret, and generate human language\cite{f0_kanbach2024genai}. At its core, NLP seeks to bridge the gap between human communication and computational understanding, making it possible for machines to process, analyze, and respond to language meaningfully\cite{f1_chavan2024opportunities,f2_schneider2024explainable,f3_voola2024leveraging}. Early NLP efforts relied heavily on rule-based systems that encoded linguistic rules and patterns manually \cite{a5_sarraf50utilizing,e1_moosavi2023effects,h1_sarraf2024deposition,f4_gadde2024all}. These systems were labor-intensive and limited in adaptability to complex or unstructured data. The advent of statistical methods in the 1990s brought significant advancements, enabling systems to learn patterns in language from large corpora of text. This era introduced probabilistic models such as Hidden Markov Models (HMMs) and Conditional Random Fields (CRFs), which became the foundation for many NLP applications like speech recognition and machine translation \cite{a4_sarraf2023chatgpt,e2_aghakhani202324,s1_sarraf2023ovitad,h2_sarraf2023reactive}.

The deep learning revolution in the 2010s marked a turning point for NLP. Neural network architectures, including recurrent neural networks (RNNs) and long short-term memory (LSTM) networks, offer the ability to model sequential data and capture dependencies over time. The introduction of attention mechanisms and the Transformer architecture, introduced in the seminal 2017 paper \textit{Attention Is All You Need} \cite{vaswani2017attention}, transformed NLP by providing a mechanism to model global relationships within text efficiently. These innovations paved the way for large-scale language models capable of handling diverse and complex linguistic tasks, leading to breakthroughs in applications such as chatbots, automatic translation, and text summarization \cite{a1_sarraf2020recent,h3_sarraf2023deposition,a3_sarraf2021comprehensive}.

Generative AI (GenAI) represents a revolutionary leap within NLP, shifting the focus from understanding and analyzing language to creating it. Powered by deep learning and large-scale Transformer-based architectures, GenAI models can generate coherent, contextually relevant, and human-like text based on input prompts. Unlike traditional NLP systems that are typically task-specific, GenAI models leverage pre-training on vast, diverse datasets followed by fine-tuning for specific use cases, making them versatile and adaptable. This paradigm has been popularized by Generative Pre-trained Transformer (GPT) models, including OpenAI’s GPT series, which have set new benchmarks in text generation, language understanding, and dialogue systems \cite{s2_sarraf2023optimal,x1_chen2024exploring, a4_sarraf2023chatgpt,e1_moosavi2023effects,e2_aghakhani202324}.

One of the key advancements in GenAI is its ability to generate text that not only adheres to grammatical and syntactical rules but also aligns with contextual and stylistic expectations \cite{s3_sarraf2023formulating,x2_haman2024using, h1_sarraf2024deposition,h2_sarraf2023reactive}. This is achieved through attention mechanisms, which allow the model to weigh the relevance of different words and phrases within a given context. Moreover, the scale of these models, with billions of parameters trained on extensive datasets, enables them to perform zero-shot and few-shot learning, where they generalize to new tasks with minimal or no additional training. This versatility has positioned GenAI as a transformative tool across industries, including healthcare, finance, education, and entertainment \cite{h3_sarraf2023deposition,s4_sarraf2020analysis,x3_masalkhi2024google,h4_sarraf2016repairing,s1_sarraf2023ovitad}.

Generative AI has opened up a vast array of capabilities that extend beyond traditional NLP applications, including:
\begin{itemize}
    \item \textbf{Text Generation:} Crafting articles, essays, stories, and creative content.
    \item \textbf{Language Translation:} Providing highly accurate translations between languages.
    \item \textbf{Text Summarization:} Generating concise summaries of extensive documents.
    \item \textbf{Conversational AI:} Powering chatbots and virtual assistants with human-like dialogue.
    \item \textbf{Code Generation:} Writing, debugging, and understanding code across programming languages.
    \item \textbf{Sentiment and Emotion Analysis:} Interpreting user sentiment and emotions from the text.
    \item \textbf{Personalization:} Delivering tailored content for marketing, e-commerce, and education.
    \item \textbf{Knowledge Retrieval:} Answering questions and retrieving information from large datasets.
    \item \textbf{Creative Problem-Solving:} Assisting in brainstorming and generating ideas for creative tasks.
    \item \textbf{Domain-Specific Expertise:} Supporting applications in specialized fields like healthcare, finance, and law.
\end{itemize}

Evaluation metrics play a crucial role in ensuring the quality and effectiveness of GenAI-generated content. The following metrics are commonly used:
\begin{itemize}
    \item \textbf{BLEU (Bilingual Evaluation Understudy):} Measures the similarity between a machine-generated text and a reference text by analyzing n-gram overlaps.
    \item \textbf{ROUGE (Recall-Oriented Understudy for Gisting Evaluation):} Focuses on recall by comparing the overlap of word sequences, particularly useful for summarization tasks.
    \item \textbf{Perplexity:} Evaluates the fluency of a language model by measuring how well it predicts a sequence of words; lower perplexity indicates better fluency.
\end{itemize}

However, these automated metrics often fail to capture deeper contextual and semantic nuances, making human evaluation indispensable. Qualitative assessments, including expert reviews and user feedback, provide richer insights into the contextual relevance and creative quality of the generated content\cite{s2_sarraf2023optimal,s3_sarraf2023formulating,s4_sarraf2020analysis,s5_sarraf2020current,x6_wei2024chatie, a2_sarraf2020binary}.

This raises an important open-ended question: Can a combination of qualitative and quantitative methods—referred to as mixed methods—be used to comprehensively assess the performance of GenAI-generated content? Addressing this question could help create a more robust and holistic evaluation framework for these transformative technologies\cite{s6_sarraf2019machine,s7_sarraf2019mcadnnet,s8_sarraf20195g,s9_sarraf2018french}.

\section{Methodological Approaches for Evaluating GenAI Output}

To comprehensively assess the performance and impact of Generative AI (GenAI) models, it is critical to explore different research methodologies. Qualitative, quantitative, and mixed-methods approaches each offer unique strengths and limitations for evaluating the content generated by GenAI. This section delves into each method, highlighting its relevance, application, and potential for enhancing the understanding of GenAI-generated outputs\cite{s11_sarraf2017eeg,s12_sarraf2017binary,s13_sarraf2016big}.

\subsection{Qualitative Research Methods}

Qualitative research focuses on exploring and understanding subjective experiences, meanings, and contexts. In the evaluation of GenAI-generated content, qualitative methods provide rich insights into the contextual relevance, creativity, and usability of the generated text. Techniques such as thematic analysis, interviews, and focus groups allow researchers to gather in-depth feedback from users or domain experts about the quality, coherence and perceived usefulness of the content. For instance, qualitative evaluations can uncover whether a generated medical summary is understandable and actionable for healthcare practitioners or patients. While qualitative methods excel at capturing nuanced feedback and identifying gaps that automated metrics might overlook, they can be time-consuming and may introduce subjectivity or bias based on the participants or evaluators\cite{s14_sarraf2016deep,s15_sarraf2016deepad}.

\subsection{Quantitative Research Methods}

Quantitative research emphasizes objective measurement and statistical analysis, making it well-suited for evaluating large-scale patterns and performance metrics. In the context of GenAI evaluation, quantitative methods often rely on automated metrics such as BLEU, ROUGE, and perplexity to assess linguistic similarity, fluency, and coherence. These metrics provide a scalable way to evaluate large datasets of generated content, enabling researchers to compare models or track improvements over iterations. Moreover, statistical analysis of user ratings or survey responses can quantify user satisfaction and engagement with the outputs. However, quantitative methods have limitations in capturing deeper contextual, cultural, or creative aspects of the content, often requiring complementary qualitative insights for a holistic evaluation\cite{s10_yang2018deep,s16_sarraf2014mathematical,s17_sarraf2014brain}.

\subsection{Mixed-Methods Research Approaches}

Mixed-methods research combines the strengths of qualitative and quantitative approaches, enabling a comprehensive evaluation of GenAI-generated content. By integrating statistical analysis with contextual and experiential feedback, mixed-methods approaches can address both the scalability of quantitative metrics and the depth of qualitative insights. For example, a mixed-methods evaluation might involve using automated metrics to filter low-quality outputs, followed by expert reviews or focus groups to assess contextual accuracy and creative quality. This dual approach is particularly valuable for complex applications, such as evaluating the effectiveness of GenAI-generated educational materials or medical summaries. While mixed-methods research requires careful planning to balance the integration of qualitative and quantitative components, it provides a robust framework for capturing the multifaceted nature of GenAI performance\cite{s10_yang2018deep}.

\section{Use Cases of Research Methods for Evaluating GenAI in Medical Imaging Manuscripts}

To demonstrate the practical application of qualitative, quantitative, and mixed-methods research in evaluating GenAI-generated content, this section presents a hypothetical use case in the field of medical imaging. A group of scientists has collaboratively written a manuscript where each section reflects the individual writing style of its contributor. The resulting manuscript lacks language coherence, readability, and overall consistency. GenAI is used to polish the original manuscript, aiming to improve its linguistic quality while preserving technical accuracy. Each research method is applied to evaluate the manuscript before and after the intervention by GenAI.

\subsection{Qualitative Research Design}

In a qualitative research design, expert reviewers, such as journal editors or senior scientists in medical imaging, are engaged to evaluate the manuscript before and after it has been polished using GenAI. These reviewers are asked to answer several open-ended questions, such as:  
1. How well does the revised manuscript achieve language coherence?  
2. Are there any improvements in readability and professional tone?  
3. Does the revised manuscript maintain the technical accuracy of the content?  
4. Are there any sections where the GenAI intervention introduced issues or over-simplifications?

After providing their responses, reviewers participate in semi-structured interviews to elaborate on their observations. Their answers and interview transcripts are analyzed using qualitative analysis software, such as NVivo or MAXQDA, to identify patterns and themes. This approach captures nuanced insights into how effectively GenAI harmonizes writing styles while preserving the scientific integrity of the content. Data analysis focuses on recurring themes like improved flow, reduced ambiguities, and potential areas where GenAI might have detracted from the manuscript's technical precision.

\subsection{Quantitative Research Design}

A quantitative approach involves using objective metrics to measure the improvement in the manuscript’s quality after being polished by GenAI. Automated tools such as Grammarly or readability indices (e.g., Flesch-Kincaid scores) are used to evaluate aspects like grammatical accuracy, sentence structure, and overall readability. Additionally, a survey with a Likert scale is distributed to a group of independent readers, including scientists and non-experts, to rate the coherence, fluency, and comprehensibility of the manuscript on a numerical scale. Statistical analysis, such as paired t-tests or ANOVA, is applied to compare the ratings and metrics before and after GenAI intervention, providing quantitative evidence of the tool’s impact.

\subsection{Mixed-Methods Research Design}

In a mixed-methods design, both qualitative and quantitative approaches are combined to provide a comprehensive evaluation of the manuscript's transformation. Automated tools are first used to assess grammatical accuracy and readability, generating quantitative metrics. These results are then supplemented by qualitative feedback from expert reviewers, who are asked to answer specific questions about the coherence, readability, and technical accuracy of the revised manuscript. Following this, reviewers are interviewed to gain deeper insights into their responses.

The qualitative data from the interviews and questionnaires are analyzed using software tools such as NVivo or MAXQDA to identify themes and patterns in the reviewers' perspectives. For instance, the software might highlight areas where reviewers consistently agree that GenAI improved readability but note concerns about potential oversimplification of technical content. The qualitative insights are then integrated with the quantitative findings to provide a more holistic evaluation, highlighting both the measurable improvements and the nuanced aspects of GenAI's performance. This integrated approach not only captures the breadth of GenAI's impact but also ensures that the evaluation reflects both objective metrics and expert judgments.
\section{Conclusion}

The rapid advancements in Generative AI (GenAI) have unlocked transformative potential across various fields, including natural language processing, content generation, and scientific writing. Evaluating the effectiveness of GenAI tools, particularly for improving the coherence, readability, and overall quality of written content, requires robust methodological frameworks. This manuscript outlined the application of qualitative, quantitative, and mixed-methods research approaches to assess the performance of GenAI in a hypothetical use case involving a medical imaging manuscript. Each approach provides unique insights into the benefits and limitations of GenAI-generated content, collectively offering a comprehensive framework for evaluating its impact.

Qualitative methods enable researchers to explore subjective aspects of GenAI performance by gathering detailed feedback from expert reviewers. Through open-ended questions and interviews, researchers can uncover nuanced insights into how effectively GenAI harmonized diverse writing styles, improved readability, and maintained technical accuracy. The use of qualitative analysis tools to process reviewer responses provides a systematic way to identify patterns and themes, ensuring that the feedback is both rigorous and actionable. This approach is particularly valuable for identifying subtleties in language use or instances where GenAI may have introduced errors or oversimplifications. By emphasizing depth and context, qualitative methods capture aspects of content quality that automated metrics might overlook.

Quantitative methods, on the other hand, offer scalability and objectivity in evaluating GenAI-generated content. Automated metrics such as BLEU, ROUGE, and readability scores provide clear, replicable measures of improvement in language coherence, fluency, and structure. Surveys with numerical ratings further quantify user satisfaction and engagement, enabling researchers to perform statistical analyses to determine the significance of improvements. While quantitative methods may not capture the intricacies of content quality, they excel at providing large-scale comparisons, making them indispensable for assessing the generalizability of GenAI improvements.

Mixed-methods approaches integrate the strengths of qualitative and quantitative research, offering a balanced framework for evaluating GenAI. By combining objective metrics with detailed expert feedback, mixed-methods research captures both the measurable improvements and the nuanced impacts of GenAI-generated content. This approach is particularly effective for evaluating complex outputs, such as scientific manuscripts, where both technical accuracy and linguistic quality are critical. Mixed-methods research ensures that evaluations are not only comprehensive but also contextually relevant, providing actionable insights for both developers and end-users.

At this stage of GenAI's evolution, these methodological approaches are crucial for quantifying the level of improvement driven by content and summary generators. GenAI tools are still subject to challenges, including biases, contextual errors, and limitations in handling highly specialized content. Rigorous evaluations help identify these limitations while highlighting areas where GenAI excels. By providing structured, evidence-based assessments, qualitative, quantitative, and mixed-methods research can guide the refinement of GenAI technologies, ensuring that they meet the diverse needs of users across disciplines.

Furthermore, these approaches allow researchers to benchmark the performance of GenAI tools against traditional writing and editing processes. This benchmarking is vital for establishing trust in GenAI systems, particularly in high-stakes domains like healthcare and scientific research. As the adoption of GenAI continues to grow, these methodologies provide the foundation for understanding its impact, ensuring that its deployment is both effective and ethical.

In conclusion, qualitative, quantitative, and mixed-methods research are indispensable for evaluating and quantifying the performance boost driven by GenAI tools. By leveraging these approaches, researchers can provide a holistic view of GenAI's capabilities, paving the way for its continued improvement and responsible application. These methodologies not only quantify the level of improvement but also ensure that GenAI tools align with the specific needs and expectations of their users, ultimately driving innovation and trust in this transformative technology.

\bibliographystyle{unsrtnat}
\bibliography{references}  






\end{document}